# Dataset Cleaning - A Cross Validation Methodology for Large Facial Datasets using Face Recognition

Viktor Varkarakis
*School of Engineeing*
*National University of Ireland Galway*
Galway, Ireland
v.varkarakis1@nuigalway.ie

Peter Corcoran
*School of Engineering*
*National University of Ireland Galway*
Galway, Ireland
peter.corcoran@nuigalway.ie

*Abstract*— In recent years, large "in the wild" face datasets have been released in an attempt to facilitate progress in tasks such as face detection, face recognition, and other tasks. Most of these datasets are acquired from webpages with automatic procedures. As a consequence, noisy data are often found. Furthermore, in these large face datasets, the annotation of identities is important as they are used for training face recognition algorithms. But due to the automatic way of gathering these datasets and due to their large size, many identities folder contain mislabeled samples which deteriorates the quality of the datasets. In this work, it is presented a semi-automatic method for cleaning the noisy large face datasets with the use of face recognition. This methodology is applied to clean the CelebA dataset and show its effectiveness. Furthermore, the list with the mislabelled samples in the CelebA dataset is made available.

*Keywords*— *face datasets, mislabeled identities, noisy samples, clean face dataset, semi-automatic cleaning, CelebA*

## I. INTRODUCTION

In the last few years, Convolutional Neural Networks (CNNs) have significantly enhanced the performance of the state-of-the-art methods in many areas, including face recognition [1]–[3]. New CNN architectures are released frequently along with new learning methodologies that push the limits of face recognition [4]–[6]. But this success is also due to the recent Big Data era that has emerged, which allows creating large face datasets with real images harvested from the Internet [7][8]. Generally, large face datasets are built in a semi-supervised way using image-search engines and thus prone to bad data samples due to mislabelling and poor image quality of some samples [9]. Often, the number of these bad data samples is not statistically significant for a particular task and they can be ignored, but, in other cases a small number of bad data samples can become quite significant and lead to poor training outcomes [10]. Therefore, cleaning the datasets from mislabelled samples is desirable for some use cases.

Consider for example a large dataset which is used to train a facial image generator, e.g. StyleGAN[11]. If the training dataset contains some mislabelled identities – i.e. wrong identity is assigned to a person – this is not critical for training a GAN that can create realistic faces as these mislabelled data samples still represent 'good' samples of facial images. However, if the task at hand switches to training a CNN to perform facial recognition, distinguishing between multiple identities, these mis-labelled samples are now 'bad' and if there are sufficient such data samples, the performance of the resulting face recognition CNN will be sub-optimal [10].

However, identifying mislabelled facial images automatically without human supervision is a very challenging task. This is due to the extreme variations of the facial images captured in the wild which can result in mis-labelling one's identity [9]. Also, it has been shown that large face datasets can typically have a noise ratio of bad data samples higher than 30% [12].

In few works, procedures and ways are described in order to minimize bad data samples when creating the dataset [10], [12]. Other researchers have tried in various ways to clean the noisy data samples from such large datasets. In [13], an anchor face that had the most neighbours was selected and a maximal subgraph starting from this anchor was regarded as the cleaning result. The authors in [14], proposed a three-stage graph-based method to clean the large face datasets using a community detection algorithm. Although these methods can clean a large part of the datasets, they have some limitations. After the cleaning procedure, the datasets may, either lack diversity as many variations are treated as outliers or the size of the dataset has decreased quite significantly due to the rigorous constraints imposed by the cleaning process.

Finally, some researchers have employed manual annotators, and have succeeded in constructing a variety of face datasets where most images are correctly labelled such as [15][16], but this approach requires significant human effort with overlapping of the data annotation to achieve a consensus on more difficult samples. It also remains prone to human error and variations in human judgement, especially on 'difficult' samples.

This work introduces a semi-automatic methodology to find and remove mislabelled samples. from large facial datasets. Such facial datasets have many practical applications in building state-of-art multimedia experiences. A methodology for improving the quality of facial data samples in such datasets is an important tool for multimedia system & content developers.

This method described here utilizes a state-of-the-art face recognition (FR) model in order to detect the outliers within of a facial dataset which is organized with multiple classes of facial identity. Based on the intra-class comparisons of the samples, the images that produce low-confidence result are considered as outliers and examined manually. This is caused by either mislabelled samples or the samples which are difficult intra-class images for the FR model. This method does not dramatically reduce the size of the original dataset or reduces the diversity of the dataset. This method has been tested on a large facial dataset and the results are presented in this paper.

In the next sections the related literature is presented, followed by a description of the cleaning methodology for facial identity datasets. Some examples of mislabelled data samples are described, and we have identified some common labelling errors across all of the dataset we have processed to date. Finally, results arising from an application of the methodology to the full CelebA dataset are presented and discussed and the list with the mislabelled samples are given in [1].

## II. RELATED LITERATURE

In the following section an overview of publicly available facial datasets used for face recognition purposes are presented. Furthermore, as a face recognition model (FR) is used in the methodology, some the state-of-the-art face recognition algorithms are described shortly.

### A. Publicly Available Facial Datasets

#### 1) CASIA-WebFace

CASIA-WebFace [17] is one of the first large public facial datasets. It contains 10.575 identities with a total of 494.414 samples. The identities belong to celebrities and are collected from the IMDb website. The size of the dataset makes it suitable for training on the face recognition task and is frequently used throughout the literature.

#### 2) CelebFaces

The CelebFaces+ dataset [18] was released in 2014 and along with the CASIA-WebFace was one of the first large publicly available datasets, as it contains 202,599 images of 10,177 identities. The dataset might also be known as CelebFaces Attributes Dataset (CelebA) [19] where the samples form CelebFaces+ are annotated with 5 landmark locations and for 40 binary attributes, providing valuable information for the researchers

#### 3) VGGFace & VGGFace2

The VGG datasets are released from the Visual Geometry Group from the University of Oxford. The VGGFace [2] dataset was released in 2015 and contains 2.6M samples from 2,622 people. VGGFace was released similarly to CASIA-WebFace mainly for training purposes. In 2018, the VGGFace2 [20] was released which consists of 3.31M samples from 9,131 celebrities. The images were downloaded from Google Image Search. The image samples from VGGFace2 cover a wider range of different ethnicities, professions and age compared to VGGFace. Furthermore, all the samples have been captured "in the wild" thus giving the dataset a desirable variation with respect to pose, lighting and occlusion conditions as well as emotions. The dataset can be used for training and testing purposes as it is divided into a train and test set. Finally, VGGFace2 provides annotations regarding the pose and the age of its samples which can be useful for researchers.

##### a) Ms-Celeb-1M

The Ms-Celeb-1M dataset [9] was created and published in 2016 by Microsoft. It is the largest publicly available face recognition dataset with over 10M samples from 100K identities. The dataset is suitable for training and testing purposes.

### B. Face Recognition Algorithms

Below a few state-of-the-art CNN based face recognition algorithms, are introduced.

#### 1) DeepFace

In 2014, Facebook published DeepFace [3]. DeepFace at the time achieved state-of-the-art accuracy (97.35%) on the famous LFW benchmark. DeepFace introduced a new alignment, employing explicit 3D face modelling in order to apply a piecewise affine transformation. Furthermore, to achieve such a performance they trained a nine-layer deep neural network with their in-house datasets which consists of 4 million face samples from more than 4,000 identities.

#### 2) FaceNet

In 2015, Google introduced FaceNet [1] and achieved accuracy of 99.63% on the LFW benchmark. FaceNet was trained on 200M images from 8 million subject. Furthermore, they introduced the triplet loss function. It requires the face triplets (an anchor, a sample of the same class as the anchor and a negative sample), and then it minimizes the distance between an anchor and a positive sample of the same identity and maximizes the distance between the anchor and a negative sample of a different identity.

#### 3) ArcFace

ArcFace [5] was published in 2018. It pushed the limits of the LFW benchmark even further as it achieved 99.83% accuracy. It also achieved state-of-the-art results on the MegaFace Challenge. Finally, the authors proposed a new loss function, additive angular margin, to learn highly discriminative features for robust face recognition

## III. METHODOLOGY

The following section describes, the methodology for finding and removing mislabelled samples in identity folders from face dataset. This methodology comprises three main stages. Initially a FR model is utilized to get an embedding from all the images. After, using the embeddings, a score for all the positive pairs from the face dataset is calculated. In the second stage the worst 2%-3% of identities of the dataset is thresholded as outliers. Finally, through a selection method, possible mislabelled samples from the thresholded identities are selected to be manually examined. The resulting data sample pairs – typically not more than a few thousand even on a large dataset - are manually examined.

### A. Scores from Face Recognition (FR) Model

Firstly, the FR model is trained (or fine-tuned) on the original dataset which is to be cleaned. Note that following an initial cleaning of the dataset, the FR can be further fine-tuned by retraining on the cleaned dataset. Thus, several iterations can be run to further improve the cleaning. This methodology leverages the power of the FR model to distinguish samples of different facial identities. The FR model must have a very good performance on the examined dataset in order to be able to detect outliers efficiently. If the FR model does not have high performance on the dataset, correctly labelled samples will be easily considered as outliers. Training / fine-tuning the FR on the dataset that will be examined, has a trade-off, as there is the possibility that the FR model will learn to classify a sample "correctly" even if it is mislabeled. Although it is assumed that the mislabeled entities comprise a very small percentage of the database and does not have a big effect on the final model. This gives the FR model the opportunity to learn the most representative embeddings during training and mislabeled samples will be treated as outliers.

---

[1] https://github.com/C3Imaging/Deep-Learning-Techniques/tree/clean-celebA

It is not a necessary step to train / fine-tune the FR on the examined dataset as FR can perform well on a dataset even if it has been trained on a different one. Although this is recommended as the FR model will thus be better optimized for the dataset that is selected to be cleaned.

Next, the embeddings for all the images are produced from the FR model. For all the positive pairs for each identity, the score is calculated using the embeddings (*pair score*). The selection of the score depends on the way the FR model was optimized (Euclidean distance, cosine similarity etc.), as models can be optimized with different losses. The proposed methodology utilizes the Euclidean distance to measure the difference between the embeddings of two images.

For each identity the scores from all the possible positive pairs are calculated and the worst score is selected as the score of the identity *(id score)*. In this way we take into consideration all the intra-class samples and it enables us to examine how good are the embeddings of the FR model produced for each identity.

*B. Outlier Selection*

After the procedure described above, each identity is assigned with an *id score.* The 2-3% of the identities with the worst *id score*, are thresholded and marked as outliers.

It is chosen to examine only the top 2-3% of the dataset as we do not want to dramatically reduce the size of the original dataset or reduces its diversity. It is desired to only to remove the most obvious outliers. There is a high possibility that the mislabelled samples are discriminated after thresholding since the FR model was not able to produce embeddings that are close enough. Although, that does not necessary means that all the samples from these identities are mislabeled.

In order to fine-grain the selection of the possible mislabeled samples from the thresholded identities, the images from the pairs that produced a low-confidence *pair score* are targeted. To do that, another threshold is defined (*pair threshold*) which is selected, based on the average value of all the *id scores* from the identities. If a pair has produced a *pair score* worse than the *pair threshold,* then the images of the pair are recorded. Also, it is noted how often an image participated in pairs that produced *a pair score,* worse than the *pair threshold*. Therefore, in this step the samples with the biggest internal embedding distance are selected as mislabeled.

To summarize, two thresholding procedures are being implemented at this step. The first one is being implemented to threshold the identities that might have mislabeled samples in their folder. The second thresholding is implemented in order to fine-grain which samples from the thresholded identities might be the mislabeled ones.

*C. Selection of Samples for Visual Examination*

As, mentioned in the previous section, the identities that might contain mislabeled samples followed by the image pairs are thresholded. Also, *image frequency* was introduced as the number of times that an image participated in a pair and had a *pair score* more than the *pair threshold.*

Based on the *image frequency* of a sample, it is determined whether it will be manually examined or not. For each identity the samples that are manually examined are selected using the following procedure. The number of pairs that have *pair score* worse than the *pair threshold* is calculated ( $NoP$ ). Then the samples are sorted in a descending order based on their *image frequency*. Starting from top to bottom a sample is selected. Every time a sample is selected, its *image frequency* is subtracted from $NoP$. Samples are selected till $NoP$ is equal or less to 0. Finally, these samples are manually examined in order to identify the mislabeled ones.

The initial experiments using the proposed methodology indicated that there are 3 common types of mislabeling in the identity folders:
   a) One main identity with 1 to $n$, mislabeled samples in the folder
   b) An identity folder with n mislabeled samples and without one of the different identities having a stronger presence than the others. By stronger presence, it is meant to have enough samples to create an identity folder (more than 4-5 samples).
   c) Two identities in the same folder.

IV. EXPERIMENTS ON CELEBA

In the next section, the methodology described is applied to clean the CelebA dataset from mislabeled samples and the result are presented with examples from each mislabeling type.

*A. Scores from FR model on CelebA*

The FR model selected for this set of experiments can be found here [2] . This is an unofficial TensorFlow implementation of FaceNet [1], built on ideas from [2]. This FR model/ implementation was selected for two main reasons. The first, being its availability and its ease of use. The second is the fact that it provides a pretrained model which reports state-of-the-art performance in the LFW test set [7]. For the purpose of this research, the available pre-trained model is fine-tuned on the CelebA dataset. The FR model's architecture is an Inception ResNetv1 [21]. The employed pretrained model that is trained with SoftMax, on the VGGFace2 dataset [20]. The input size of the network is an $160x160$ image and the output is a 512-embeding.

---

[2] https://github.com/davidsandberg/facenet

The reason for the fine-tuning is for the FR to be more dataset specific and have a higher performance on the dataset that will be examined. In the fine-tuning process, the same configurations as in training were used, with a reduced learning rate. For more information regarding the training of the FR model and data preparation see [2].

In Fig.1 the ROCs on the CelebA dataset is presented for the models before and after fine-tuning. The ROC curves, shows that the FR model after fine-tuning on the CelebA, performs better than the pretrained model. Therefore, it will point to the identity folders that may have mislabeled samples more effectively. This is because the performance is increased, resulting in a lower false positive error. This also illustrates the need for the FR model to be trained / fine-tuned on the examined dataset.

After fine-tuning is completed, the 512-embedding for all the samples of CelebA dataset are calculated. The score used for this set of experiments is the Euclidean distance between the embeddings. For each identity, the scores from all the possible positive pairs are calculated and the worst score (in this case the highest Euclidean distance) is selected as the score of the identity *(id score)*.

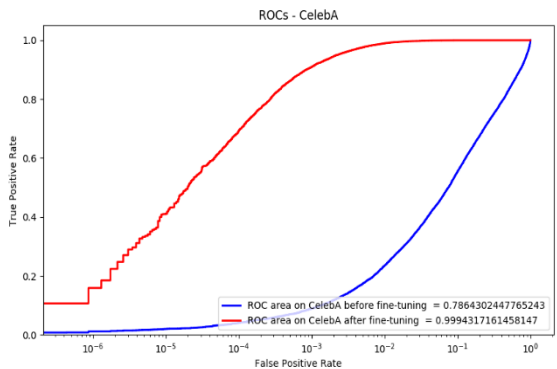

Fig. 1. The ROC curves on the CelebA dataset. The blue line, shows the performance for FR model before fine-tuning and the red for the FR model after-finetuning.

### B. Identifying possible mislabeled identity folders

As mentioned in the section B of the Methodology, in two thresholding procedures take place at this step.

In the first thresholding procedure, the identities with the worst 3% *id score* (in this case with highest Euclidean distance) are thresholded. These identities are considered as outliers, as they might contain mislabeled samples in their folders. This means 310 identities.

Afterwards, a second thresholding takes place. This is implemented in order to fine-grain the selection of the possible mislabeled samples that may exist in the thresholded identity folders (section III-B).

First, the *pair threshold* is calculated as described in the Methodology (section B) which is the average of all the *id scores* from all the identities and is equal to 1.

Therefore, for the thresholded identities, all the positive image pairs that have Euclidean distance more than 1 (*pair threshold*), are recorded. Also it is noted how often an image exist in pair with *pair score,* worse than the *pair threshold,* and is defined as *image frequency*. In Fig.2 illustrates the distribution of the *id score* from all the identities.

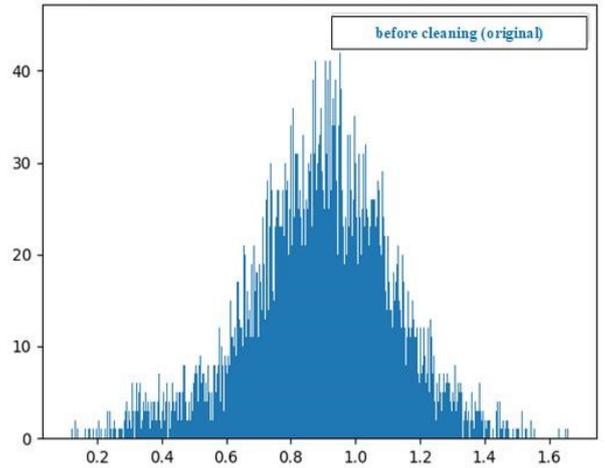

Fig. 2. The distribution of the *id score* from all the identities from the CelebA dataset before applying the method proposed for cleaning the dataset.

### C. Results

After the two thresholding operations (initially for the identities and afterwards for its samples), the *image frequency* is finally calculated. As mention in the section C, of the Methodology, *the image frequency* is used to determine which samples will be manually examined for being possible mislabeling.

Applying the methodology, the three types of mislabeling in the identity folders appeared as well as cases where the methodology flagged an identity folder which by examining its samples, it did not have any mislabeling, but it contained samples with high variation. Below some examples are presented for each case, along with the different actions that were chosen for cleaning the dataset, depending on the mislabeling type.

1. Two identities in one folder

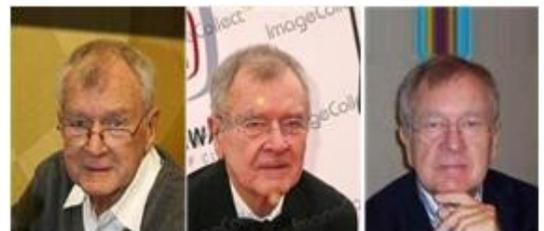

(a)

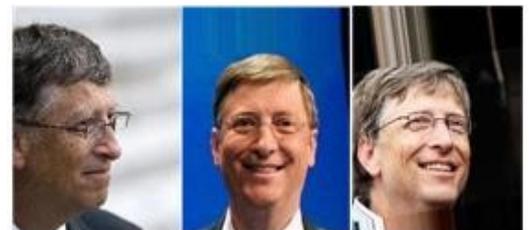

(b)

Fig. 3. Example of a mislabeled folder in CelebA, which contains two different identities.

In this case the mislabeling type of having two identities in the same folder was detected using the described technique. Fig. 3a and Fig. 3b are samples, obviously from 2 different identities but found on the same identity folder. In this cases, one identity is retained, and the samples of the other identity are removed

2. One identity with $n$ mislabeled samples

In this case the technique used, flagged an identity folder which contained samples from one identity but also some mislabeled samples. Fig. 4a, shows the mislabeled samples existing in the identity folder. Fig. 4b shows the samples belonging to the same identity.

In these cases, the mislabeled samples were deleted. In case that the mislabeled samples were many and the main identity was left with less than 2-3 samples, the folder was as well removed.

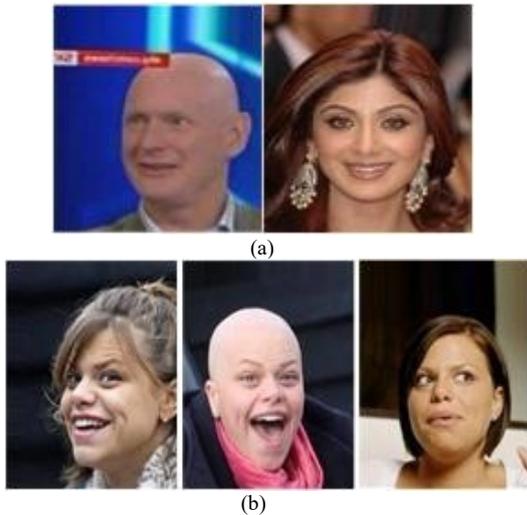

Fig. 4. Example of an folder in CelebA, which contains one main identity with 1 to n, mislabeled samples in the folder.

3. An identity folder with $n$ mislabeled samples

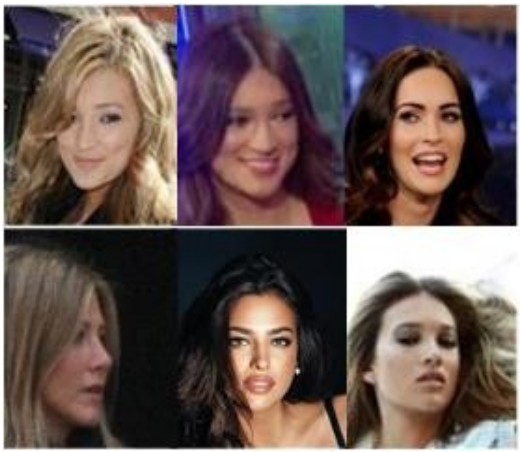

Fig. 5. Example of an folder in CelebA, with n mislabeled samples and without one of the different identities having a stronger presence than the others.

In this example a folder with $n$ different identities were detected, as it can be seen in Fig.5. In this type of mislabeling, where a folder had $n$ mislabeled samples without one identity having a stronger present (By stronger presence, it is meant to have enough samples to create an identity folder (more than 4-5 samples)), the whole identity folder was deleted.

4. High variation in an identity folder

Finally, there were cases were the methodology indicated an identity folder as an outlier. Though by examining its samples, it was detected that the samples belong to the same identity. The technique indicated this identity folder as an outlier due to its high variation. In Fig. 6, such an example is illustrated. In case a folder was considered an outlier without having any mislabeled samples, no further action was taken, and its samples were retained.

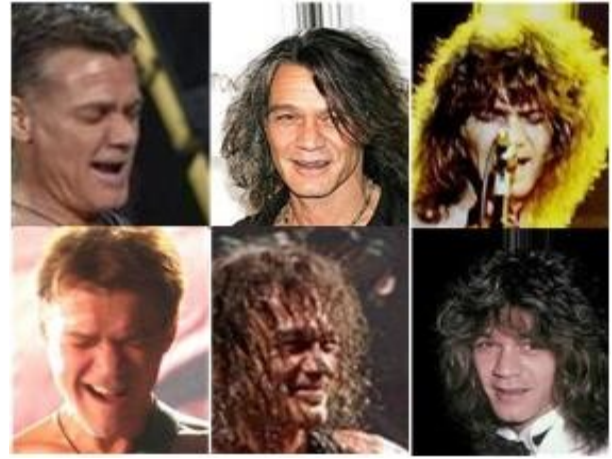

Fig. 6. Example of an folder in CelebA, shows a case where the methodology was unsuccessful. As it identified this folder as outlier with possible mislabeled samples but the folder just contained difficult intra-class samples due to variation for the FR model.

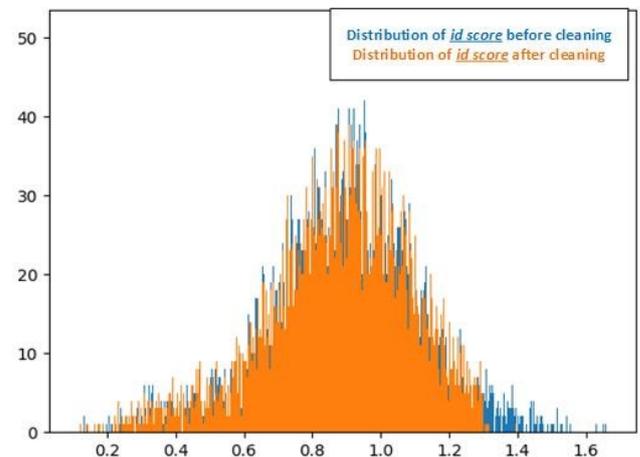

Fig. 7. The distribution of the *id score* from all the identities from the CelebA dataset before (blue) applying the method proposed for cleaning the dataset and after (orange). The distribution of the *id scores* was reduced when the mislabeled samples were removed, showing the effect of cleaning in the dataset with the proposed method.

In total, from the 310 identities that were flagged using the proposed method only 9 of them did not have any mislabeling samples. The other 301 identities selected contained mislabeled samples. In Fig. 2, the distribution of the *id score* for all identities is shown before cleaning the CelebA dataset. In Fig. 7 the distribution of the *id score* is shown after the cleaning of the CelebA dataset, in comparison with the initial distribution of the *id score* from Fig2. It can be seen from the Fig.7, that majority of the high scores were due to the mislabeled samples, as after removing the mislabeled samples the distribution of the *id score* was reduced.

The CelebA dataset as mentioned earlier, consist of 202,599 images from 10,177 identities. After applying the methodology for finding and removing mislabeled samples as described earlier, it remains with 197,477 samples from 9,996 identities. The list with the mislabeled samples is publicly available in [1].

## V. Discussion and Future Work

In the sections above, a (semi-automatic) technique for identifying and removing mislabeled samples in terms of identity is described. The technique utilizes a face recognition model trained / fine-tuned on the examined dataset in order to discover outliers in an identity folder that shall be examined as it is possible to contain mislabeled face samples. This methodology was applied to clean the CelebA dataset and the results are presented in section IV-C. In addition, the list with the mislabeled samples can be found in [1]. This technique can be applied to any face dataset annotated with identities in order to "clean" it so that the dataset can be used with more certainty as a considerable number of mislabeled samples will be eliminated.

In this preliminary work our main goal has been to demonstrate the effectiveness of the methodology to provide a minimal curation of the dataset. In other words, we seek to retain as many of the original data samples as possible to ensure that the diversity of the original dataset is preserved. There is still a lot of work to apply these techniques across additional large datasets and to further automate the methodology and develop additional analysis tools and quality metrics to fully demonstrate its capability to improve the quality of these datasets.

Also, this technique will be examined in order to observe the influence and how to achieve the best configuration for setting the identity and pair threshold, as the tuning of this threshold has not been explored in detail in this preliminary work. Furthermore, this methodology will be used to identify mislabeled samples in other face datasets. The methodology should also be compared with some datasets that have been manually cleaned and we are currently signing some license agreements to gain access to a number of such 'clean' datasets. It is expected that some comparisons can be provided for presentation at QoMEx 2020.

Finally, it would be useful to automate additional aspects of the cleaning process and approaches to reduce the computational complexity of the methodology. A number of these will be explored, working in collaboration with other researchers later this year.

## Acknowledgment

This research is funded under the SFI Strategic Partnership Program by Science Foundation Ireland (SFI) and FotoNation Ltd. Project ID:13/SPP/I2868 on Next Generation Imaging for Smartphone and Embedded.